\documentclass[sigconf, nonacm]{acmart}

\AtBeginDocument{%
  }

\usepackage{graphicx}
\usepackage{amsmath}
\usepackage{multirow}
\usepackage{subfigure}
\usepackage{enumitem}
\usepackage{tikz}
\usepackage{float}
\usepackage{soul}
\sethlcolor{black}

\newcommand*\circled[1]{\tikz[baseline=(char.base)]{
            \node[shape=circle,draw,inner sep=0.6pt] (char) {#1};}}

\usepackage{xcolor}
\definecolor{sampledot}{RGB}{0,59,238}
\definecolor{bpdl}{RGB}{0,167,238}
\definecolor{cdl}{RGB}{0,167,70}

\newcommand{\logo}{\includegraphics[height=1em]{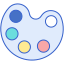}}
\newcommand{\modelname}{\textsc{PaLeTTe}}
\newcommand{\modelnamex}{\textsc{PaLeTTe }}
\newcommand{\synthmap}{\textsc{SynthMap+}}
\newcommand{\synthmapx}{\textsc{SynthMap+ }}
\newcommand{\synthtext}{\textsc{SynthText}}
\newcommand{\synthtextx}{\textsc{SynthText }}

\begin{document}

\title{Hyper-Local Deformable Transformers for Text Spotting on Historical Maps}

\author[*]{Yijun Lin}
\authornote{Corresponding author}
\affiliation{%
  \institution{University of Minnesota, Twin Cities}
  \city{Minneapolis}
  \state{MN}
  \country{USA}
}
\email{lin00786@umn.edu}

\author{Yao-Yi Chiang}
\affiliation{%
  \institution{University of Minnesota, Twin Cities}
  \city{Minneapolis}
  \state{MN}
  \country{USA}
}
\email{yaoyi@umn.edu}

\renewcommand{\shortauthors}{Yijun Lin \& Yao-Yi Chiang}

\begin{abstract}
Text on historical maps contains valuable information providing georeferenced historical, political, and cultural contexts. However, text extraction from historical maps has been challenging due to the lack of (1) effective methods and (2) training data. Previous approaches use ad-hoc steps tailored to only specific map styles. Recent machine learning-based text spotters (e.g., for scene images) have the potential to solve these challenges because of their flexibility in supporting various types of text instances. However, these methods remain challenges in extracting precise image features for predicting every sub-component (boundary points and characters) in a text instance. This is critical because map text can be lengthy and highly rotated with complex backgrounds, posing difficulties in detecting relevant image features from a rough text region. This paper proposes \modelname, an end-to-end text spotter for scanned historical maps of a wide variety. \modelnamex introduces a novel hyper-local sampling module to explicitly learn localized image features around the target boundary points and characters of a text instance for detection and recognition. \modelnamex also enables hyper-local positional embeddings to learn spatial interactions between boundary points and characters within and across text instances. 
In addition, this paper presents a novel approach to automatically generate synthetic map images, \synthmap, for training text spotters for historical maps. The experiment shows that \modelnamex with \synthmapx outperforms SOTA text spotters on two new benchmark datasets of historical maps, particularly for long and angled text. 
We have deployed \modelnamex with \synthmapx to process over 60,000 maps in the David Rumsey Historical Map collection and generated over 100 million text labels to support map searching.
The project is released at \url{https://github.com/kartta-foundation/mapkurator-palette-doc}
\end{abstract}



\maketitle

\section{Introduction}

Detecting and recognizing text from scanned documents is essential for effectively managing them and facilitating analysis and accessibility to their content. This is particularly important for historical and cultural documents, such as historical manuscripts, books, building \& infrastructure plans, engineering drawings, handwritten records, and maps. Historical maps, for example, offer valuable information about how territories, landscapes, and civilizations have evolved over time, which are extremely crucial resources required in various scientific domains, including humanity, geography, urban planning, cartography, and environmental science~\cite{chiang2014survey, chiang2020using, chiang2023geoai}. 

This paper focuses on text spotting on scanned historical maps, an important data source having wide varieties (e.g., in cartographic designs and scanned quality) but lacking (1)~\textit{effective approaches}, (2)~\textit{text labels for training}, and (3)~\textit{benchmark datasets}. 
Text on historical maps, including names of places, labels for physical features (e.g., building types), and descriptions of areas or landmarks, provides unique insights on georeferenced historical, political, and cultural contexts~\cite{chiang2020using, chiang2014survey}. For example, text on historical geological maps indicates locations of rock formations, faults, folds, and mineral deposits, for which digital versions of this information are not available at the scale and coverage of the historical maps.\footnote{See the recent AI for Critical Mineral Assessment Competition from DARPA and USGS: \url{https://criticalminerals.darpa.mil/}.}

Numerous types of scanned historical maps are accessible online, such as the US Geological Survey topographic maps~\cite{davis1893topographic}, the Sanborn maps~\cite{lamb1961sanborn}, the Ordnance Survey maps~\cite{hewitt2011map}, and a vast range of maps in the David Rumsey Historical Map Collection~\cite{rumsey2008david}. Accessing desired scanned historical maps typically requires searching through map metadata, which are often limited or unavailable. 
Generating comprehensive metadata for individual map scans requires expert knowledge and extensive manual work. Therefore, accurately and automatically extracting text from scanned historical maps is crucial for enhancing their searchability and facilitating the generation of map metadata~\cite{chiang2023geoai}.

Previous attempts to extract map text involve ad-hoc steps to tailor and finetune existing text detection and recognition approaches for a specific map style~\cite{arundel2022deep, chiang-text-recognition, li2018intelligent} or rely on external place name databases~\cite{li2020automatic, weinman2019deep}. 
Recent text spotting approaches have shown promising results on benchmark datasets (mostly scene images), indicating their potential for historical maps. One category of these methods requires a careful design of the region of interest (RoI) operation to extract image features within the detected regions for recognition~\cite{he2017mask, liu2018fots, liu2020abcnet, liu2021abcnet2}. 
Another category needs additional post-processing to group the results, such as segmentation-based approaches~\cite{huang2022swintextspotter, liao2020mask, wang2021pgnet}. These methods are either poor in recognition due to imprecise detected regions or difficult to generalize for text instances with wide varieties of curvatures, sizes, spacings, orientations, and placement, e.g., historical map text. 

\begin{figure}[htbp]
  \centering 
    \subfigure[TESTR's sampling locations ($\textcolor{sampledot}\bullet$) are around the text center ($\textcolor{red}\bullet$) for predicting the first character (left) and the last character (right)]{
    \begin{minipage}[t]{\linewidth}
        \centering 
        \includegraphics[width=\linewidth]{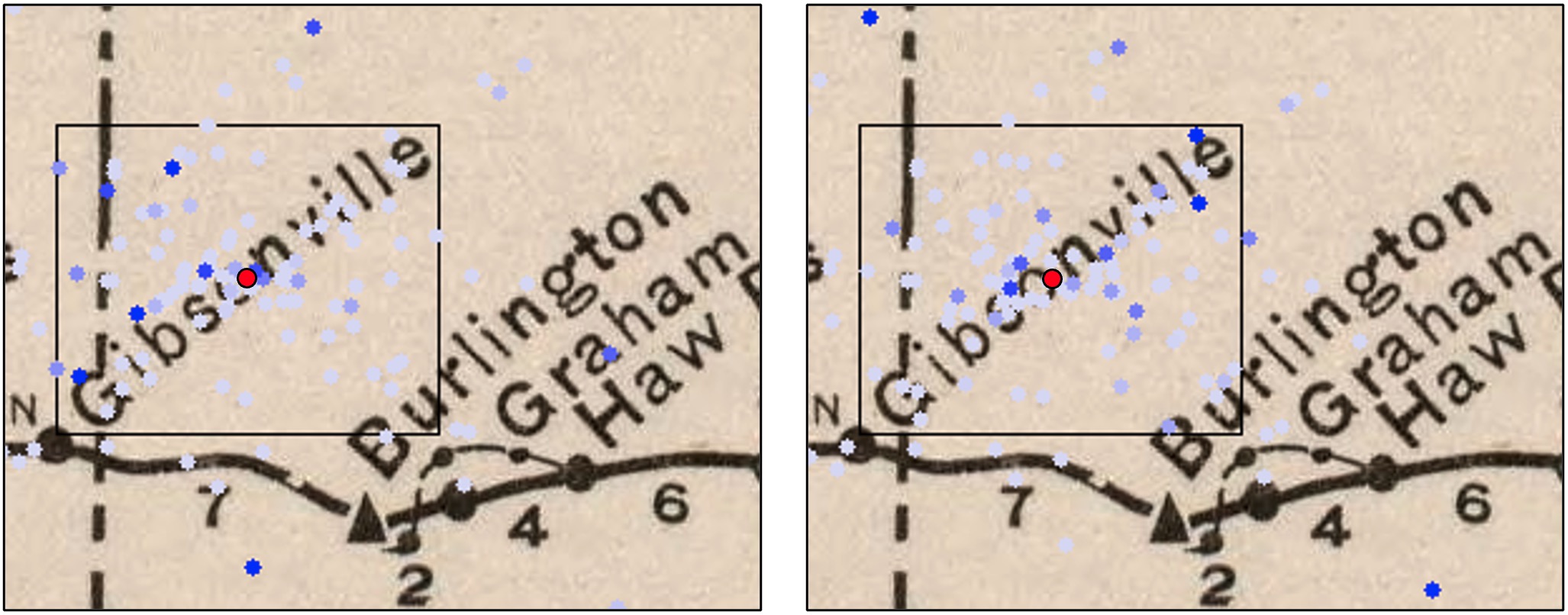}   \vspace{-0.05in}
        \label{figure: attend1} 
    \end{minipage}} 
    \subfigure[\modelnamex uses character locations ($\textcolor{red}\bullet$) to sample hyper-local features ($\textcolor{sampledot}\bullet$) for predicting the first character (left) and the last character (right)]{
    \begin{minipage}[t]{\linewidth}
        \centering
        \includegraphics[width=\linewidth]{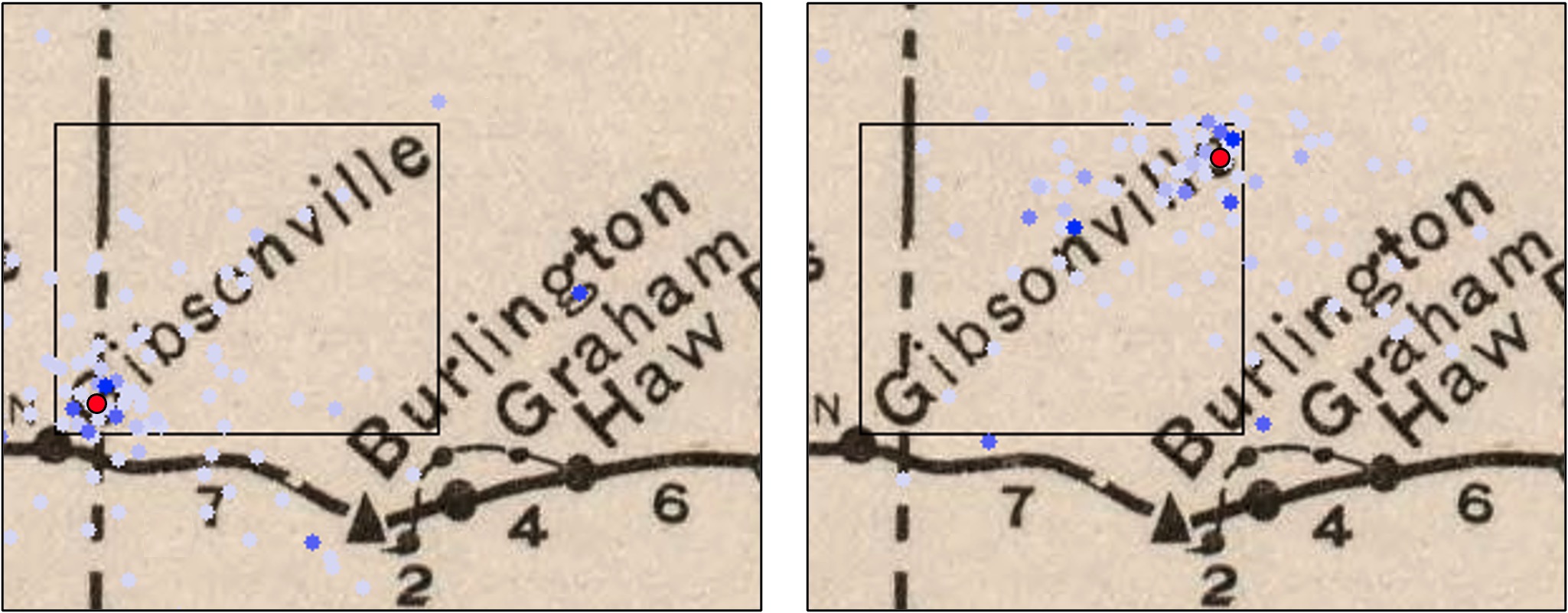}   \vspace{-0.05in}
        \label{figure: attend2}
    \end{minipage}} \vspace{-0.1in}
  \caption{Comparison between reference points ($\textcolor{red}\bullet$) and sampling locations ($\textcolor{sampledot}\bullet$) in TESTR and \modelnamex when querying the first character (left figure) and the last character (right figure) . Darker blue dots have higher attention scores.} \vspace{-0.1in}
  \label{figure: attend} 
\end{figure}

The third category mitigates these issues by detecting and recognizing text in parallel, for example, utilizing a deformable DETR~\cite{zhu2020deformable} with a separate character decoder on top of the boundary point decoder~\cite{zhang2022text, ye2023deepsolo}. However, there are two remaining challenges in handling lengthy, curved, and highly rotated text on historical maps. \textbf{First}, deformable DETR and its variant spotting models learn image features for detection and recognition around a coarse reference point that roughly indicates the position of a text instance (e.g., the instance center). This works well for object detection, for which the networks predict one object category for a detected object. However, text spotting requires further prediction on multiple sub-components (boundary points and characters) of a detected object (text instance). The reference point(s) and, consequentially, the sampled image features can be misaligned with the target boundary points or characters, especially for text instances that are long, winding, or with extra-large character spacing. For example, Figure~\ref{figure: attend1} shows that the deformable DETR-based text spotter, TESTR, samples features from a fixed location (the text center as $\textcolor{red}\bullet$) that can be distant from the target characters. \textbf{Second}, all boundary points and characters of a text instance share the same position information (e.g., the text center coordinate), which cannot well capture the inter- and intra- spatial relations between these sub-components within and across text instances.

Furthermore, training these machine learning models requires a large number of text labels, while creating text annotations requires extensive manual work. Although there are existing synthetic datasets~\cite{wang2012end, jaderberg2014synthetic} that are inexpensive and scalable alternatives and provide ground-truth annotations, they are typically scene images that differ significantly from historical maps in terms of text and background styles. Thus, the lack of text labels for training is still a challenge for text spotting on historical maps.

This paper presents \modelname\footnote{\modelnamex refers to hy\textbf{P}er-loc\textbf{AL} d\textbf{E}formable \textbf{T}ransformers for \textbf{TE}xt spotting} \logo~with \synthmapx for extracting text from historical maps. \modelnamex is an end-to-end text spotter for historical maps with text instances of arbitrary shapes and can be long \& highly-rotated. The main idea of \modelnamex is to explicitly model the individual boundary points and characters of a text instance as the target objects for spotting. \modelnamex progressively refines the position of boundary points and character centers and uses them as (1) the new reference points to sample image features around the target sub-components, i.e., hyper-local sampling (see Figrue~\ref{figure: attend2}), and (2) position information to describe the inter- and intra- spatial relations between sub-components, i.e., hyper-local positional embeddings. 
In addition, we propose \synthmap, a synthetic dataset of historical-styled map images to facilitate training text spotters for historical maps.\footnote{The synthetic dataset is released at \url{https://zenodo.org/records/14480731}} \synthmapx follows cartographic rules to place text labels of various styles and integrates with the background extracted from real historical maps.

We have deployed \modelnamex with \synthmapx in the mapKurator system~\cite{kim2023mapkurator}, a complete pipeline for extracting and linking text from historical maps. The system has processed over 60,000 maps in the David Rumsey Historical Map collection and the extracted text labels have been incorporated into the metadata platform to support map searching by Luna Imaging.\footnote{Search by Text-on-Maps: \url{https://mailchi.mp/stanford/apr2023-ai-advancements-in-map-studies}}
The main contributions of the paper include the following,
\begin{itemize}[leftmargin=*]
\item We propose \modelnamex that introduces (1) hyper-local sampling to learn precise image features for predicting boundary points and characters, and (2) hyper-local position embeddings to capture spatial relations between boundary points and characters within and across text instances. 
\item We propose an approach that automatically generates \synthmapx of various styles with text labels for training text spotters for historical maps. Additionally, we introduce a new, highly diversified benchmark dataset for evaluating text spotting performance on real historical maps.
\item \modelnamex with \synthmapx is the first text spotter capable of handling a wide variety of scanned historical maps, thus opening up a vast asset of valuable information in tens of thousands of historical maps.
\end{itemize}

\section{\modelname}
\begin{figure}[ht]
  \centering 
  \vspace{-0.05in}
  \includegraphics[width=\linewidth]{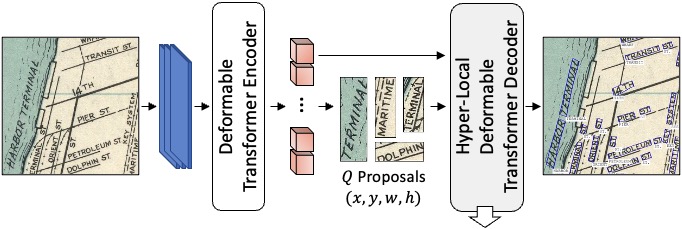}  
  \includegraphics[width=\linewidth]{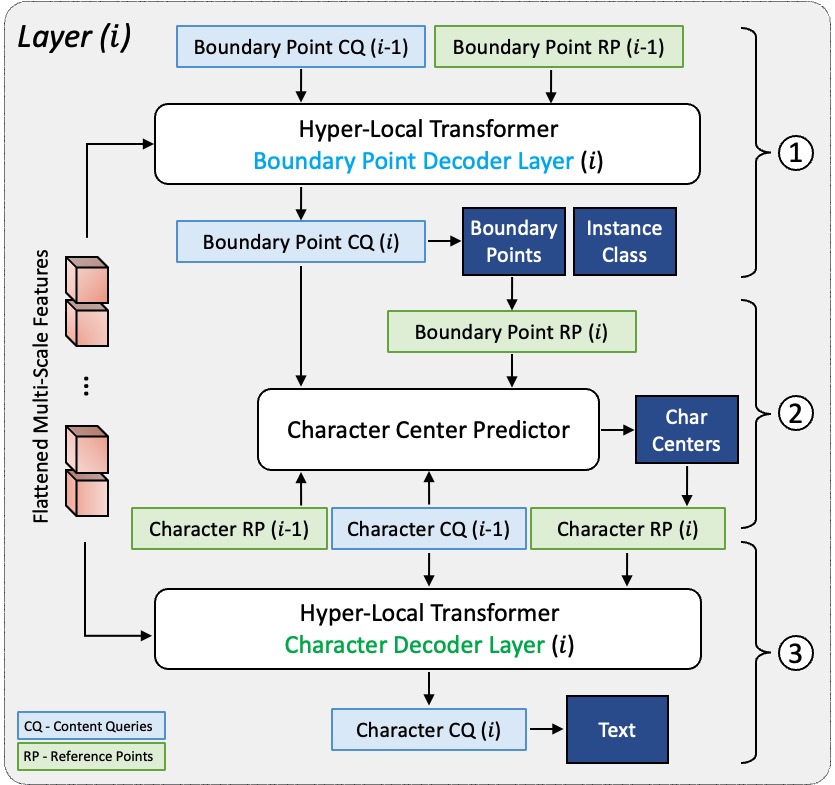}  
  \caption{The network architecture of \modelnamex (top). The Hyper-Local Deformable Transformer decoder contains multiple layers (bottom). \modelnamex progressively refines the content queries of boundary points and characters, and their reference points layer by layer.} 
  \label{figure: archtecture}   \vspace{-0.15in}
\end{figure}

Figure~\ref{figure: archtecture} shows the network architecture of \modelname. \modelnamex uses a CNN backbone followed by a Deformable Transformer encoder to extract multi-scale feature maps from the input image. \modelnamex uses these image features to identify the top $Q$ box proposals indicating the approximate locations of text instances. 
In the decoder, the first layer uses the proposal centers as the initial reference points to sample the image features for predicting boundary points. Then, a character center predictor takes the detected boundary points and character information (initialized as zeros) as input to predict character centers, serving as the reference points for recognizing characters. The rest of the layers iteratively update the position of the boundary points and character centers as the new reference points to sample localized image features for text detection and recognition.

\subsection{Deformable Attention in Text Spotting}
DETR~\cite{carion2020end} is known for its high accuracy in object detection tasks. However, DETR computes the attention weight between every pair of features in the entire feature maps, which can include feature information that is not related to the target query and is computationally expensive. Deformable DETR~\cite{zhu2020deformable} addresses this issue by using a data-dependent sparse attention mechanism that aims to sample only relevant features for attention. Specifically, given a set of $L$-level multi-scale feature maps~\cite{he2016deep}, denoted as ${\{\mathbf{x}_l\}}^L_{l=1}$, where each level $\mathbf{x}_l \in \mathbb{R}^{C \times H_l \times W_l}$. Let $\mathbf{p}(q)$ be the normalized coordinates of the reference point to sample image features for a content query~$q$. The multi-scale deformable attention works as,
\begin{equation}
\begin{aligned}
& \text{MSDeformAttn}(q, \mathbf{p}(q), {\{\mathbf{x}_l\}}^L_{l=1}) = \\
&    \sum^H_{h=1} W_h \{ \sum^L_{l=1} \sum^K_{k=1} A_{hlk}(q) \cdot W'_h \mathbf{x}_l [\phi_l(\mathbf{p}(q)) + \Delta \mathbf{p}_{hlk}(q)] \}
\label{eq1}
\end{aligned}
\end{equation}
where $h$, $l$, $k$ refer to the indices of the attention head, the feature level, and the sample points. $A_{hlk}$ denotes the attention weight for query $q$, normalized across $L \times K$ sample points. The function $\phi_l$ rescales the normalized coordinates $\mathbf{p}(q)$ to the $l$-th level feature map and $\Delta \mathbf{p}$ computes the sample offset from $\mathbf{p}(q)$ for the content query, and their sum decides the sampling locations.

Using Deformable DETR for text spotting~\cite{zhang2022text, ye2023deepsolo} typically starts with a proposal generator to generate top $Q$ box proposals, represented as $\{(x, y, w, h)_i \}^Q_{i=1}$, where $(x,y)$ are the center coordinates and $(w,h)$ are the box scale. The centers of these proposals serve as reference points in the decoder. The decoder samples image features from reference points with learned offsets and refines the content queries to predict boundary points and characters layer by layer.


\subsection{Hyper-Local Deformable Transformer}
For a target instance, sharing the reference point (e.g., using the proposal center) across all content queries has a limitation in that the reference point indicates the coarse position of a text instance and is not closely related to its sub-components, i.e., the $N$ boundary points and $M$ characters.\footnote{We use $*_n$ as the symbols related to boundary points and $*_m$ for characters.} This will lead to imprecise position information and sample locations for extracting image features. To overcome this issue, we propose \textit{hyper-local sampling} that generates features from localized reference points and \textit{hyper-local positional embedding} that indicates the precise location for each sub-component of the instance.

For a proposal $i$, \modelnamex initializes its reference points for the boundary point content queries $( q_n^{(i,1)}, ..., q_n^{(i,N)} )$ using the proposal center coordinates $(x, y)$. The positional embeddings of the boundary point content queries are also initially identical, $\varphi((x, y))$, where $\varphi$ is a sinusoidal positional encoding function. The $N$ boundary point composite queries for instance~$i$ are formulated as,
\begin{displaymath}
    \{\hat{q}_n^{(i, j)}\}^N_{j=1} = \{q_n^{(i,j)} + \varphi((x, y)^{(i)})\}^N_{j=1}
\end{displaymath} 

In Figure~\ref{figure: archtecture} \circled{1}, the \textcolor{bpdl}{boundary point decoder layer} leverages deformable attention to learn feature maps from sampled image features around the reference points (Eq.~\ref{eq1}). The output represents the updated information about the boundary points by looking at the multi-scale image features. \modelnamex further refines the output by adding the positional embeddings again and conducting self-attention between the boundary points within an instance and between instances. This step enables the boundary points to adjust their content in accordance with each other. 

The final output becomes the new boundary point content queries for predicting the boundary points, ${(\Bar{x}_n^{(j)}, \Bar{y}_n^{(j)})}^N_{j=1}$. Then \modelnamex updates the reference points using the predicted boundary points, aiming to move the sampling locations close to the target boundary points for hyper-local sampling in the next layers. In addition, \modelnamex computes the new positional embeddings from the predicted boundary points, enabling hyper-local position information. The new composite queries are the addition of the updated content queries and the new positional embeddings:
\begin{displaymath}
    \{\hat{q}_n^{(i,j)}\}^N_{j=1} = \{q_n^{(i,j)} + \varphi(  (\Bar{x}_n^{(i,j)}, \Bar{y}_n^{(i,j)} )  )\}^N_{j=1}
\end{displaymath}
In this way, \modelnamex progressively refines the boundary points, consequentially, the reference points and positional embeddings towards the target boundary points layer by layer to learn localized and precise image features for text detection.

\modelnamex also applies the hyper-local strategy to the \textcolor{cdl}{character decoder layer} in Figure~\ref{figure: archtecture} \circled{3}. To retrieve localized character information, \modelnamex predicts the center of each character as the reference point, indicating the precise position of the characters. Specifically, for a proposal~$i$, \modelnamex computes the composite queries for $M$ characters by adding the character content queries, ($q_m^{(i, 1)}, ..., q_m^{(i, M)}$), and their positional embeddings of the character centers, ${(\Bar{x}_m^{(i,j)}, \Bar{y}_m^{(i,j)})}^M_{j=1}$. Then \modelnamex samples hyper-local image features around the character centers for queries using deformation attention.  \modelnamex further refines the output by adding the positional embeddings of the character center locations and adopting self-attention between the characters within a text instance. This enables \modelnamex to capture the relationships between characters based on their content and locations. The output is used to update the character content queries and predict characters.
Here, \modelnamex introduces a new module to predict character centers using boundary points and character information (Section~\ref{sec: ccp}). 

The hyper-local strategy has two main advantages: (1) each content query has its own explicit positional prior, which allows for sampling localized image features around target sub-components (boundary points and characters) of a text instance. This strategy enables extracting high-quality image features and results in superior training convergence since it localizes the sampling region; (2) the positional queries encode the locations of individual reference points for sub-components, which benefits the intra-class attention between boundary points/characters within the text instance and the inter-class attention across text instances.

\subsection{Character Center Predictor}~\label{sec: ccp}
In Figure~\ref{figure: archtecture} \circled{2}, we propose a character center predictor to generate the centers and update the character reference points accordingly, providing precise guidance on the sampling locations in the character decoder layer. Typically, the center of a character is highly related to the character content and the text instance boundary. Leveraging this assumption, the predictor takes character content queries and boundary point content queries with their respective position information to predict character centers.

For a text instance $i$,\footnote{Here we drop $(i)$ in $q^{(i, j)}$ for simplicity.} the predictor uses the character content queries $\{ q_m^{(j)} \}_{j=1}^M$ with their positional embeddings $\{ \mathbf{p}(q_m^{(j)}) \}_{j=1}^M $ as new queries ($\mathcal{Q}$), and the boundary point content queries $\{ q_n^{(j)} \}_{j=1}^N$ with their positional embeddings $\{ \mathbf{p}(q_n^{(j)}) \}_{j=1}^N $ to form key and value sets ($\mathcal{K}$, $\mathcal{V}$), and predicts character centers as follows, 
\begin{displaymath}
\begin{aligned}
    \mathcal{Q} &= \big\{\mathbf{w_q} \cdot (q_m^{(j)} + \mathbf{p}(q_m^{(j)})) \big\}_{j=1}^{M}  \\
    \mathcal{K} &= \big\{\mathbf{w_k} \cdot (q_n^{(k)} + \mathbf{p}(q_n^{(k)})) \big\}_{k=1}^{N}  \\
    \mathcal{V} &= \big\{\mathbf{w_v} \cdot  q_n^{(k)} \big\}_{k=1}^{N} \\
    \mathcal{A} &= \big\{\mathcal{Q}_j \cdot \mathcal{K}_k \big\}_{j=1,\ k=1}^{M,\ N} \\
    C_{coords} &= \big\{MLP( \sum_{k=1}^N \mathcal{A}_{j, k} \cdot \mathcal{V}_k    ) \big\}_{j=1}^{M} \\ 
\end{aligned}
\end{displaymath}
where $\mathbf{w_q}$, $\mathbf{w_k}$, and $\mathbf{w_v}$ are linear projections. $\mathcal{A}$ is the attention map containing $N \times M$ elements, and $C_{coords}$ are the character center coordinates predicted using cross attention between characters and boundary points followed by a 3-layer $MLP$. 

This module not only predicts the character centers to support hyper-local sampling and positional embeddings, but also bridges the gap between two decoders and enables the collaboration between the boundary points and characters of a text instance. The predicted centers will be supervised by the truth character centers during training. If a text instance is shorter than the predefined maximum text length, we set the rest centers as the middle of the last point of the top curve and the first point of the bottom curve, i.e., the center line tail. We discuss the details in Section~\ref{sec:it}.


\subsection{Optimization}
\subsubsection{Bipartite Matching} After obtaining the prediction set $\hat{Y}$, \modelnamex uses the Hungarian algorithm \cite{kuhn1955hungarian} to achieve an optimal matching between $\hat{Y}$ and the ground truth set $Y$ that minimizes the matching cost $\mathcal{C}$: \vspace{-0.05in}
\begin{displaymath}
    \arg \min_{\varphi} \sum^G_{g=1} \mathcal{C}(Y^{(g)}, \hat{Y}^{(\varphi(g))}) 
\end{displaymath}
where $G$ is the number of ground truth and $\varphi(g)$ is the predicted instance index matched to the $g$-th ground truth instance. 

Regarding the cost $\mathcal{C}$, previous work~\cite{zhang2022text} uses the boundary point class and position similarity. However, pairs with similar boundary point positions could increase optimization difficulties~\cite{ye2022deepsolo}. DeepSolo~\cite{ye2022deepsolo} introduces character similarity in the cost. However, character recognition performance at the early training stage might be much poorer than detection, potentially affecting the matching process.  
\modelnamex adds the comparison between the predicted character centers with the ground truth centers, making it robust in the matching process. Thus, for the $g$-th ground truth instance and a target instance, the cost function is:
\begin{displaymath}
\begin{aligned}
    \mathcal{C} = \lambda_{cls} FL(\hat{b}^{(\varphi(g))}) & + \lambda_{coord} \sum^{N-1}_{n=0} \lVert p^{(g)}_n -  \hat{p}^{(\varphi(g))}_n \rVert \\
    & + \frac{\lambda_{center}}{|C|} \sum^{C-1}_{c=0} \lVert p^{(g)}_c -  \hat{p}^{(\varphi(g))}_c \rVert  \\
\end{aligned}
\end{displaymath}
where $\lambda_{cls}$, $\lambda_{coord}$, $\lambda_{center}$ are the hyper-parameters, $\hat{b}^{(\varphi(g))}$ is the probability for the text-only instance class~\cite{zhu2020deformable}, $p_n$ and $p_c$ are the coordinates of $n$-th boundary point and $c$-th character center, respectively, and $|C|$ is the number of characters.  

\subsubsection{Overall Loss} \modelnamex uses the same encoder losses for proposals as~\cite{zhang2022text, ye2022deepsolo, ye2023deepsolo}, denoted as $\mathcal{L}_{enc}$.
For the $j$-th query in the decoder, the loss function is:
\begin{displaymath}
    \mathcal{L}_{dec}^{(j)} = \lambda_{cls}\mathcal{L}^{(j)}_{cls} + \lambda_{coord}\mathcal{L}^{(j)}_{coord} + \lambda_{ct}\mathcal{L}^{(j)}_{ct} + \lambda_{char}\mathcal{L}^{(j)}_{char}
\end{displaymath}
where $\mathcal{L}^{(i)}_{cls}$ and $\mathcal{L}^{(i)}_{coord}$ are the classification loss and boundary points loss, $\mathcal{L}^{(j)}_{ct}$ is the $L1$ loss for character centers, and $\mathcal{L}^{(j)}_{char}$ is the $CE$ loss for character recognition. We also introduce the intermediate decoder loss to regularize the output of each decoder layer. Thus, the overall loss $\mathcal{L} = \mathcal{L}_{enc} + \sum_j \mathcal{L}_{dec}^{(j)}$.

\subsection{Iterative Training}~\label{sec:it}
Although character centers are available or can be easily obtained in synthetic datasets, retrieving character centers in human annotations is non-trivial. Inspired by~\cite{xing2019convolutional}, we propose an iterative training method for \modelnamex to mitigate the requirement for character centers in using human-annotated data. During pretraining, we leverage massive synthetic data with low-cost character center information and a small number of human annotations to train \modelnamex in a weakly supervised manner, i.e., \modelnamex does not compute the character center loss if centers are unavailable. Now, the pretrained \modelnamex has the capability to predict character centers using the character center predictor.

\begin{figure}[ht]
  \centering 
  \includegraphics[width=0.9\linewidth]{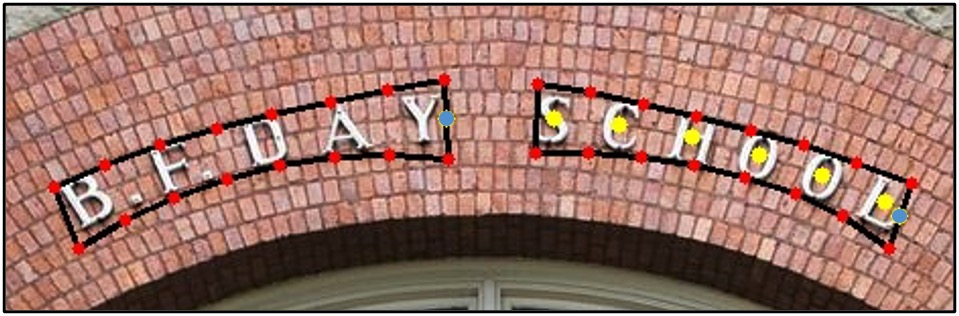} 
  \caption{An example of training \modelnamex in a weakly supervised way. \modelnamex iteratively adds the correctly predicted character centers (e.g., $\textcolor{yellow}\bullet$) for training. Black polylines are true text boundaries; red dots are true boundary points; yellow dots are predicted character centers; and blue dots are fake centers for the rest empty characters.} \vspace{-0.1in}
  \label{figure: center_example}  
\end{figure}

During finetuning, we iteratively add the ``correct'' character center predictions into training samples and use these centers for training. Figure~\ref{figure: center_example} provides an example to illustrate how \modelnamex gradually identifies the ``correct'' character centers for real-world images. If all center locations of the characters in a word (i.e., yellow dots) are within the ground truth text boundary (i.e., black polylines), we consider the text instance having a ``correct'' character center prediction (e.g., ``SCHOOL''). Otherwise, we only keep the fake centers (i.e., blue dots) for the rest empty characters (e.g., ``B.F.DAY''). Then, we add these correctly predicted centers to the training data. After the first round of finetuning, we apply the finetuned model on the images again to retrieve new ``correct'' character centers. We repeat this process until the number of ``correct'' character center predictions is stable. This iterative training strategy on \modelnamex alleviates the requirement on expensive human-annotated character centers, instead leveraging the predicted centers and training the model in a weakly supervised manner.  

\section{\synthmapx}

Manual annotations on real-world images (e.g., scene images and scanned documents) are prohibitively expensive. Therefore, existing text spotters leverage synthetic datasets for training. However, existing synthetic datasets~\cite{gupta2016synthetic, liu2020abcnet} mainly focus on scene images and do not work well for historical maps because map text can have various cartographic styles, character spacing sizes, and rotation angles. For example, map text typically follows the corresponding geographic features (e.g., rivers) in a complex line-styled background (e.g., roads, contour lines). \citet{li2021synthetic} propose to generate synthetic map images for text detection on historical maps. However, the dataset lacks text transcriptions and features only one map style (i.e., Ordnance Survey), limiting its adaptability to train text spotting models for diverse map styles in practice.

\begin{figure}[htbp!]
  \centering 
  \includegraphics[width=\linewidth]{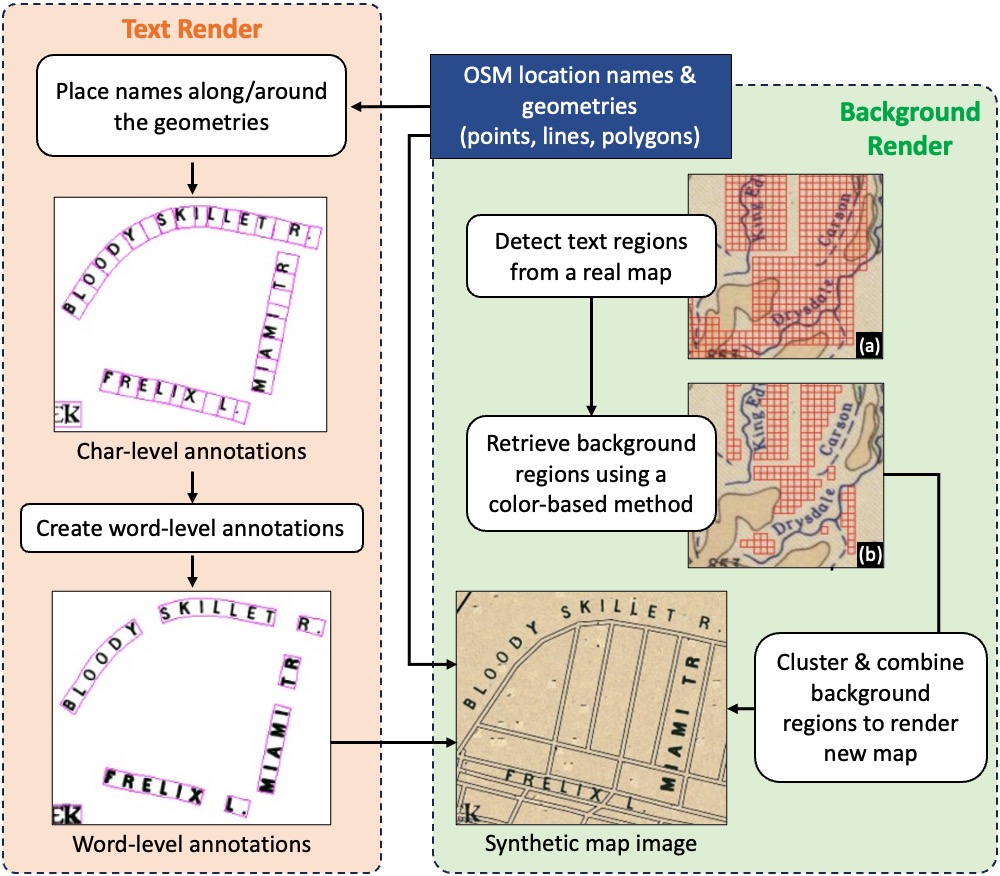} 
  \caption{The workflow of generating synthetic map images and text labels, including word-level boundary points, character centers, and transcription.} 
  \label{figure: synmap_generation}  
\end{figure}

We propose a novel approach that automatically generates map images of historical styles along with text annotations, named \synthmap, to train text spotters for historical maps. Figure~\ref{figure: synmap_generation} shows the workflow for generating \synthmapx images and text labels. We first collect a large number of location names, geographic features (lines and polygons\footnote{We exclude point features because they do not provide diverse geometry shapes.}), and their types (e.g., rivers and industrial areas) from OpenStreetMap\footnote{https://www.openstreetmap.org/}. Our approach contains two modules, \textcolor{orange}{text render} and \textcolor{cdl}{background render}, to produce synthetic map images of a wide variety. 

The text render (TR) uses the QGIS label placement API~\cite{lawhead2017qgis} to draw location names according to the shapes of geographic features on an empty canvas. For line features, the text instances follow the line curvature (e.g., roads and railways). For polygon features, the text instances follow the contour of the polygon boundary (e.g., lakes). TR randomly sets text font, size, letter spacing, and word spacing for text labels, providing various styles for placement. Then, TR retrieves the bounding box of each character and groups them to produce the boundary polygon, character centers, and transcription for each text instance.

\begin{figure}[htbp!]
  \centering  
  \includegraphics[width=\linewidth]{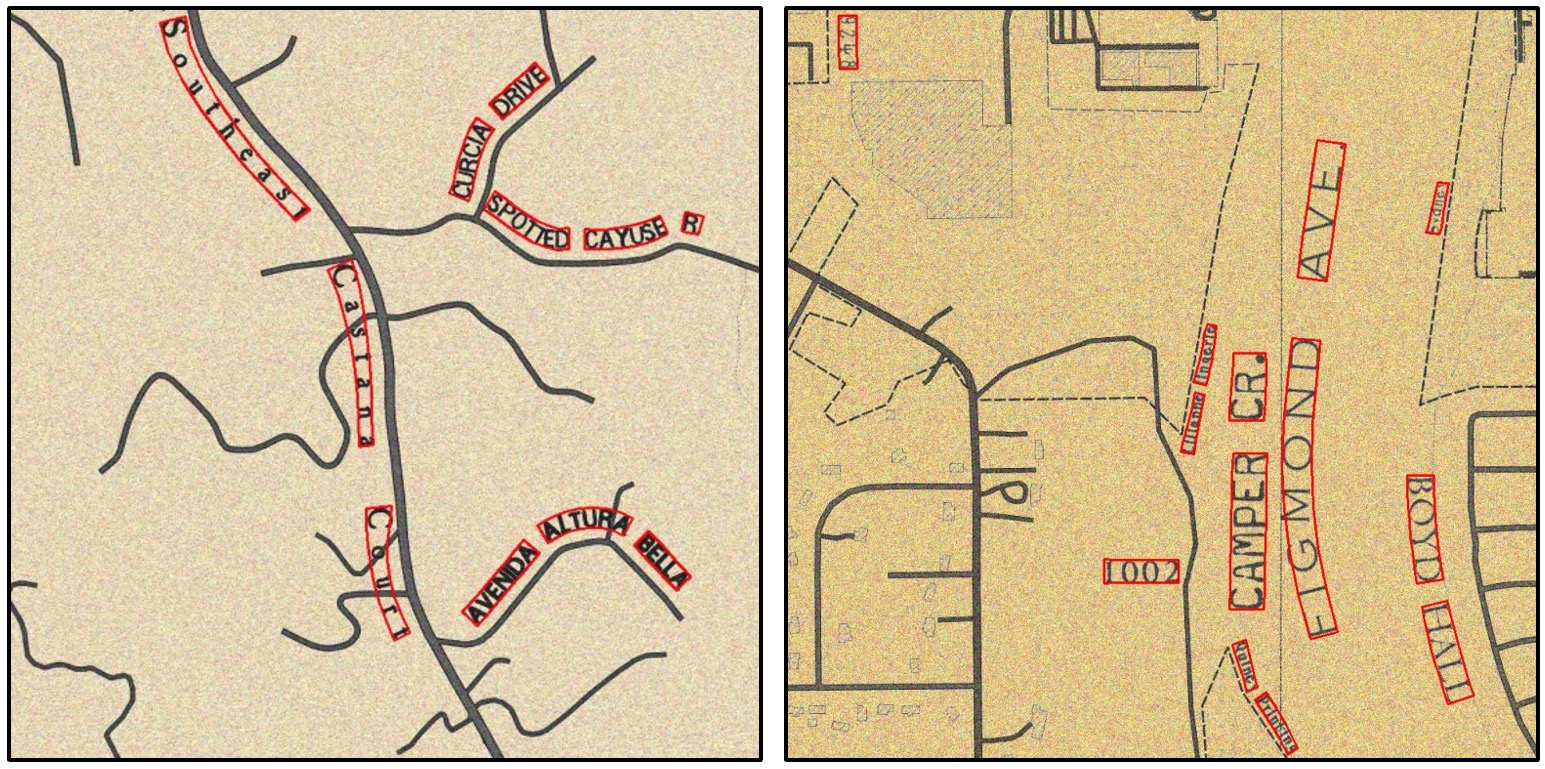}  \vspace{-0.2in}
  \caption{Examples of \synthmapx images. Red boundaries are word-level annotations.} 
  \label{figure: synmap}
  \vspace{-0.1in}
\end{figure}

The background render (BR) aims to create realistic scanned map backgrounds in a wide variety. The idea is to identify representative background regions from real historical maps and use them as new map backgrounds. Specifically, BR first detects text regions from maps in the David Rumsey Historical Map Collection and creates a rectangle buffer around the detection results to retrieve pixels around text. Instead of using pixels, BR constructs grid cells of 8$\times$8 pixels as a unit to preserve the local pattern of the texture and color. The red cells in Figure~\ref{figure: synmap_generation} \textcolor{white}{\hl{(a)}} show the extracted text regions. BR adopts K-means to classify these cells into background and foreground classes based on their colors, see Figure~\ref{figure: synmap_generation} \textcolor{white}{\hl{(b)}}. Then BR uses a second K-means to group these background cells into clusters as the representative map style profiles and use them to render new maps. When creating a synthetic map image, BR randomly selects one map style and its background cells will be reconstructed into an image, serving as the color and texture. Our approach also uses QGIS API to paint geographic features of various types using predefined cartographic rules on these backgrounds. Finally, our approach merges the output from TR and BR to create synthetic map images. Figure~\ref{figure: synmap} shows two examples in \synthmap.

\section{Experiments and Results}
\subsection{Training Datasets}
\subsubsection{\synthtext} 
Following~\cite{gupta2016synthetic, liu2020abcnet}, we create a new synthetic dataset that contains character centers extracted from character bounding polygons. We add background images of ancient canvas styles from WikiArt~\cite{artgan2018} and the OSM location names for text rendering. \synthtextx consists of 84,182 and 48,000 images of mostly straight and curved text, respectively. 
\subsubsection{\synthmap} We create a synthetic map dataset containing a total of 73,657 images with 1,102,940 annotations. 
\subsubsection{External Human-Annotated Datasets} We add human annotations of existing real-world scene images to facilitate training: ICDAR15~\cite{karatzas2015icdar}, Total-Text~\cite{ch2020total}, and MLT~\cite{nayef2017icdar2017}. 


\subsection{Evaluation Map Datasets}
We evaluate the text detection and recognition performance on two manually annotated historical map datasets, following the standard evaluation protocols~\cite{liu2020abcnet}. 

\subsubsection{Grinnell-UMass-31}
\citet{ray2018historical} annotated 31 scanned historical maps from 9 atlases in the David Rumsey Map Collection. We crop the maps into patches of 1,000$\times$1,000 pixels and select 500 images with the largest number of text instances. The text instances truncated at the image edge or having all numeric characters are excluded during the evaluation (i.e., treated as ``DON'T CARE''). The final dataset contains 15,143 non-numeric text instances.

\subsubsection{Rumsey-309}
We create a large and diversified historical map benchmark dataset by randomly selecting and annotating 309 image patches (1,000$\times$1,000 pixels) from 54 diverse historical maps in the David Rumsey Historical Map Collection. We follow the annotation guideline~\cite{asante2017data} and generate ground truth annotations. The final dataset contains 13,145 non-numeric text labels. 

\subsection{Experimental Settings}
Following the convention for training text spotters, we pretrain \modelnamex (and baselines) with synthetic and human-annotated datasets and finetune models with Total-Text~\cite{ch2020total}. We choose Total-Text because (1) we do not have enough historical map images for both training and testing, and (2) we find that existing text spotters finetuned with Total-Text usually perform the best on our evaluation datasets. We provide details of network architecture, hyper-parameters, and training settings in Appendix~\ref{appendix: b}.

\subsection{Results and Discussion}
\subsubsection{Overall Performance} We compare \modelnamex with several SOTA spotting models: ABCNet-v2~\cite{liu2021abcnet2}, SWINTS~\cite{huang2022swintextspotter}, TESTR~\cite{zhang2022text}, and DeepSolo~\cite{ye2023deepsolo}. Table~\ref{table: res1} reports the spotting performance of pretrained and finetuned models. 
Pretrained \modelnamex outperforms baselines with E2E (no lexicon) F-score gains of $\geq$3.3\% on \textit{Grinnell-UMass-31} and $\geq$5.8\% on \textit{Rumsey-309}. These results highlight the effectiveness of \modelnamex in historical map text spotting. Additionally, finetuning models with Total-Text generally enhances the performance. The improvement on \textit{Grinnell-UMass-31} is less significant compared to \textit{Rumsey-309}. This might be because \textit{Grinnell-UMass-31} contains only nine map styles, making finetuning on diverse scene images less impactful. Also, finetuned \modelnamex models achieve less improvement compared to the pretrained ones because, even with iterative training, there are some text instances that \modelnamex cannot correctly locate their character centers, limiting its training ability. The overall results demonstrate the robustness of \modelname. We further explore the model performance on example challenging cases, including lengthy and highly rotated text instances.


\begin{table*}[htbp]
\centering
\renewcommand{\arraystretch}{0.9}
\caption{Spotting performance of \textbf{pretrained} and \textbf{finetuned} models on \textit{Grinnell-UMass-31} and \textit{Rumsey-309}. ``None'' refer to no lexicon. ``Full'' uses the lexicon of all words in the test set. The best results are in \textbf{bold}, and the second best are \underline{underlined}.} \vspace{-0.05in}
  \begin{tabular}{c | l | c c l l c | c c l l c} \toprule
    && \multicolumn{5}{c|}{\textit{Grinnell-UMass-31}} & \multicolumn{5}{c}{\textit{Rumsey-309}} \\ \midrule
    & \multirow{2}{*}{Method} & \multicolumn{3}{c}{Detection\;\;\;\;\;\;\;\;\;\;\;\;\;\;\;\;} & \multicolumn{2}{c|}{E2E} & \multicolumn{3}{c}{Detection\;\;\;\;\;\;\;\;\;\;\;\;\;\;\;\;} & \multicolumn{2}{c}{E2E}\\
    && P & R & F & None & Full & P & R & F & None & Full \\ \midrule 
    \multirow{6}{*}{Pretrained} 
    & ABCNet & 82.48 & 81.35 & 81.91 & 54.46 & 66.50 & 80.16 & 78.36 & 79.25 & 44.83 & 57.49 \\
    & SWINTS    & \underline{88.90} & 72.46 & 79.89 & 58.14 & 68.95 & \underline{86.93} & 71.52 & 78.48 & 56.97 & 68.31  \\
    & TESTR     & 87.53 & \underline{82.30} & \underline{84.84}  & \underline{68.05} & \underline{77.94} & 85.72 & \underline{79.07} & \underline{82.26} & \underline{65.99} & \underline{75.59} \\
    & DeepSolo  & 87.15 & 74.57 & 80.37 & 67.54 & 74.69 & 78.13 & 68.1 & 72.77 & 60.11 & 67.27  \\ \cmidrule{2-12}
    & \modelname & \textbf{89.20} & \textbf{83.09} & \textbf{86.04} ($\uparrow$1.4\%) & \textbf{70.27} ($\uparrow$3.3\%) & \textbf{79.13} & \textbf{88.01} & \textbf{81.77} & \textbf{84.77} ($\uparrow$3.1\%) & \textbf{69.82} ($\uparrow$5.8\%) & \textbf{78.75} \\ \midrule
    \multirow{6}{*}{Finetuned} 
    & ABCNet & 84.46 & 81.56 & 82.98 & 53.84 & 65.03 & 87.59 & 83.63 & 85.56 & 45.91 & 58.44 \\
    & SWINTS    & 88.38 & 74.29 & 80.73 & 59.40 & 68.33 & \underline{91.92} & 77.28 & 83.97 & 60.45 & 71.50 \\
    & TESTR     & 89.27 & \textbf{83.40} & \underline{86.24} & \underline{69.95} & \underline{78.09} & 92.25 & \textbf{86.36} & \underline{89.21} & \underline{73.44} & \underline{81.54} \\ 
    & DeepSolo  & \underline{89.44} & 76.02 & 82.18 & 67.92 & 75.24 & 89.18 & 78.11 & 83.28 & 68.52 & 76.26  \\ \cmidrule{2-12}
    & \modelname & \textbf{89.98} & \underline{83.53} & \textbf{86.64} ($\uparrow$0.5\%) & \textbf{71.67} ($\uparrow$2.5\%) & \textbf{78.94} & \textbf{93.56} & \underline{87.28} & \textbf{90.31} ($\uparrow$1.2\%) & \textbf{75.87} ($\uparrow$3.3\%) & \textbf{83.63} \\ \bottomrule 
  \end{tabular}
  \label{table: res1} \vspace{-0.05in}
\end{table*}

\subsubsection{Text Length} 
Table~\ref{table: res2} presents the E2E recall on text instances of lengths larger than 7 and 10. \modelnamex outperforms baselines with significant enhancements, e.g., 8\% on \textit{Grinnell-UMass-31} and 2\% on \textit{Rumsey-309} when text length $\geq 10$. The results demonstrate that the use of a single fixed reference point or sampling reference points from the center line is insufficient for long text instances compared to the hyper-local strategy in \modelnamex.

\begin{table}[htbp]
\centering
\renewcommand{\arraystretch}{0.9} 
\caption{E2E recall of \textbf{pretrained} models on the text length ($l$) : $l \geq 7$ and $l \geq 10$} \vspace{-0.05in}
  \begin{tabular}{l | c c | c c } \midrule
    & \multicolumn{2}{c|}{\textit{Grinnell-UMass-31}} & \multicolumn{2}{c}{\textit{Rumsey-309}}  \\ 
     & $l \geq 7$ & $l \geq 10$ & $l \geq 7$ & $l \geq 10$  \\ \midrule 
    \# Text    & 4242  & 826  & 2436  & 393 \\ \midrule 
    ABCNet-v2  & 57.02 & 49.93 & 43.22 & 36.41 \\
    SWINTS     & 57.22 & 42.63 & 52.20 & 39.94 \\
    TESTR      & 71.31 & \underline{62.98} & \underline{66.14} & \underline{58.26} \\
    DeepSolo   & \underline{72.51} & 62.36 & 63.21 & 55.64 \\ \midrule
    \modelname & \textbf{74.74} & \textbf{68.08} & \textbf{68.04} & \textbf{59.29} \\ \midrule
  \end{tabular}
  \label{table: res2} 
\end{table}

\subsubsection{Text Orientation}
Table~\ref{table: res3} shows the E2E recall on the text instances of rotation angles in [30$^\circ$, 60$^\circ$) and [60$^\circ$, 90$^\circ$]. The text instances within these ranges constitute approximately 25\% of the datasets, and \textit{Rumsey-309} has 11\% more instances with large angles than \textit{Grinnell-UMass-31}. The baselines (ABCNet-v2 and DeepSolo) that rely on Bezier curves as boundary representations perform worse than those directly using polygon points (TESTR and \modelname) when spotting text with high rotation angles, especially [60$^\circ$, 90$^\circ$]. \modelnamex outperforms baselines on both datasets, especially having an improvement of 4.8\% in [60$^\circ$, 90$^\circ$] on \textit{Rumsey-309}, which further demonstrates the effectiveness of the hyper-local sampling strategy in \modelnamex.

\begin{table}[htbp]
\centering
\renewcommand{\arraystretch}{0.9} 
\caption{E2E recall of \textbf{pretrained} models on two ranges of text orientation degrees: [30$^\circ$, 60$^\circ$) and [60$^\circ$, 90$^\circ$]} \vspace{-0.05in}
  \begin{tabular}{l | c c | c c } \midrule
     & \multicolumn{2}{c|}{\textit{Grinnell-UMass-31}} & \multicolumn{2}{c}{\textit{Rumsey-309}}  \\ 
     & [30, 60) & [60, 90] & (30, 60] & (60, 90] \\ \midrule 
    \# Text    & 1873  & 1275  & 1981  & 1534 \\ \midrule 
    ABCNet-v2  & 22.67 & 15.94 & 20.51 & 9.84\\ 
    SWINTS     & 40.69 & 37.12 & 45.50 & 47.06 \\
    TESTR      & \underline{52.65} & \underline{46.18} & \underline{57.98} & \underline{53.38} \\
    DeepSolo   & 46.22 & 36.14 & 50.97 & 44.84 \\ \midrule
    \modelname & \textbf{54.74} & \textbf{47.17} & \textbf{62.58} & \textbf{55.22} \\ \midrule
  \end{tabular}
  \label{table: res3} 
\end{table}

\subsubsection{Effect of \synthmapx}
Table~\ref{table: res4} shows that introducing \synthmapx significantly improves performance in both \modelnamex and the baseline models. \synthmapx enables text spotters to train from a wide variety of map styles in text and background, boosting their performance compared to only using \synthtext. For example, incorporating \synthmapx during pretraining leads to a remarkable 24.8\% increase in E2E performance for \modelname. We show the results for other baselines in Appendix~\ref{appendix: c}.

\begin{table}[htbp]
\centering
\renewcommand{\arraystretch}{0.9} 
\caption{Effect of \synthmap. Models are pretrained with: \synthtextx ($\mathcal{S}$), human-annotations ($\mathcal{H}$) (ICDAR, Total-Text, MLT), and \synthmapx ($\mathcal{M}$). The spotting performance are reported on \textit{Rumsey-309}.} \vspace{-0.05in}
  \begin{tabular}{c | c | c c c c} \midrule
                           & Datasets & Det-P & Det-R & Det-F & E2E-None \\ \midrule
    \multirow{3}{*}{TESTR} & $\mathcal{S}$ + $\mathcal{H}$ & 77.1 & 75.7 & 76.4 & 53.0 \\ 
                           & $\mathcal{S}$ + $\mathcal{M}$ & 84.0 & 73.9 & 78.6 & 61.0 \\ 
                           & $\mathcal{S}$ + $\mathcal{H}$ + $\mathcal{M}$ & 85.7 & \underline{79.0} & \underline{82.2} & \underline{65.9} \\ \midrule
    \multirow{3}{*}{\modelname} & $\mathcal{S}$ + $\mathcal{H}$ & 79.2 & 77.1 & 78.1 & 55.9 \\ 
                           & $\mathcal{S}$ + $\mathcal{M}$ &  \underline{86.7} & 76.9 & 81.5 & 64.6 \\ 
                           & $\mathcal{S}$ + $\mathcal{H}$ + $\mathcal{M}$ & \textbf{88.0} & \textbf{81.7} & \textbf{84.7} & \textbf{69.8} \\ \midrule
  \end{tabular}
  \label{table: res4} 
\end{table} 


\subsubsection{Ablation Studies} 
We conduct ablation studies to further analyze (a) the effect of hyper-local sampling, (b) the effect of hyper-local positional embeddings, and (c) the effect of incorporating character center distance in the cost of bipartite matching. For (a), we remove hyper-local sampling for detecting boundary points (\modelname-$\textsc{wo-HLD}$) and recognizing character (\modelname-$\textsc{wo-HLR}$) respectively by replacing the dynamic reference points with the fix proposal center. For (b), we replace the hyper-local positional embeddings with the positional embedding of the center point (\modelname-$\textsc{wo-HLPE}$). For (c), we omit the character center distance in the cost of bipartite matching (\modelname-$\textsc{wo-B}$). Table~\ref{table: res5} presents several insights: (a) without hyper-local sampling results in a 2.9\% reduction in detection F1 (first row) and 5.1\% in recognition performance (second row). This finding highlights the effectiveness of using boundary points and character centers as the respective reference points to sample localized image features for the spotting tasks. (b) \modelnamex exhibits a marginal decrease in detection and recognition performance when hyper-local positional embeddings are excluded, demonstrating the necessity of using localized position information for predicting the target sub-components. (c) incorporating character center distance in bipartite matching leads to a slight improvement, indicating that the character center distance enables a robust bipartite matching between the predictions and ground truth by considering the character positions.

\begin{table}[htbp]
\centering
\renewcommand{\arraystretch}{0.9} 
\caption{Ablation Studies of \modelnamex on \textit{Rumsey-309}} \vspace{-0.1in}
  \begin{tabular}{l | c c c c } \midrule
     & Det-P & Det-R & Det-F & E2E-None  \\ \midrule
    \modelname-$\textsc{wo-HLD}$  & 86.3 & 79.1 & 82.3 & 67.9  \\ 
    \modelname-$\textsc{wo-HLR}$  & 86.9 & 80.9 & 83.8 & 66.4 \\
    \modelname-$\textsc{wo-HLPE}$ & 87.4 & 78.9 & 83.0 & 68.3  \\ 
    \modelname-$\textsc{wo-B}$  & 87.9 & 80.0 & 83.2 & 69.1 \\ \midrule
    \modelname    & 88.0 & 81.7 & 84.7 & 69.8  \\ \midrule
  \end{tabular} 
  \label{table: res5} \vspace{-0.15in}
\end{table}

\subsubsection{Visual Analysis}
\begin{figure}[htbp]
    \centering 
    \includegraphics[width=\linewidth]{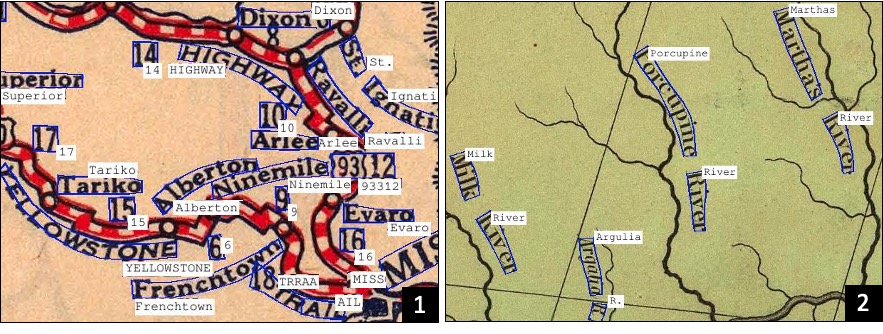}  
    \includegraphics[width=\linewidth]{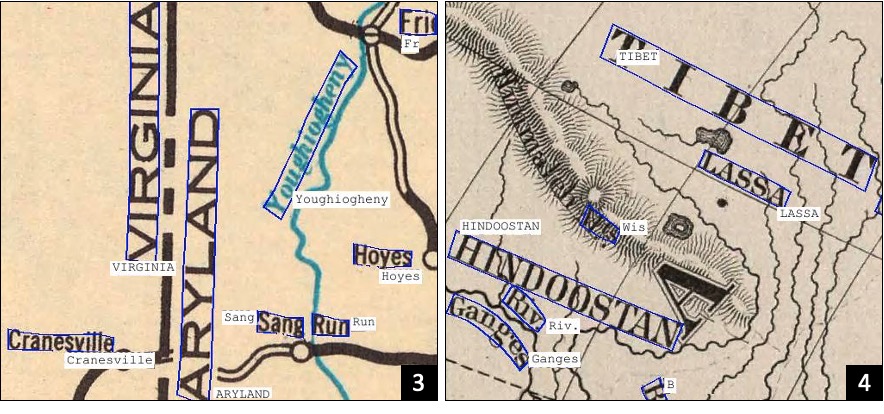}  
    \includegraphics[width=\linewidth]{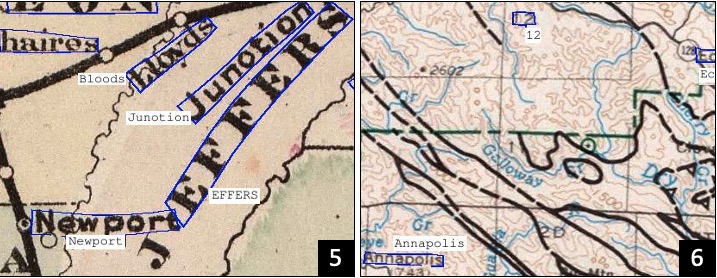}  
    \caption{Visualization of spotting results}
  \label{figure: res}  
\end{figure}

Figure~\ref{figure: res} presents several example results showing that \modelnamex can accurately detect and recognize complex text instance arrangements and styles. 
However, \modelnamex faces difficulties when the text size is extremely large and characters in a text instance are far away, e.g., the character ``A'' in map image 4, when some other text instances or linear features overlap the target text instance (e.g., the character ``J'' is separated from ``JEFFERS'' by ``Newport''), and when many linear features exist on/around a text instance. We believe that more diverse types of synthetic training data could help alleviate this problem. 

\section{Deployment}
\begin{figure}[H]
    \centering 
    \includegraphics[width=\linewidth]{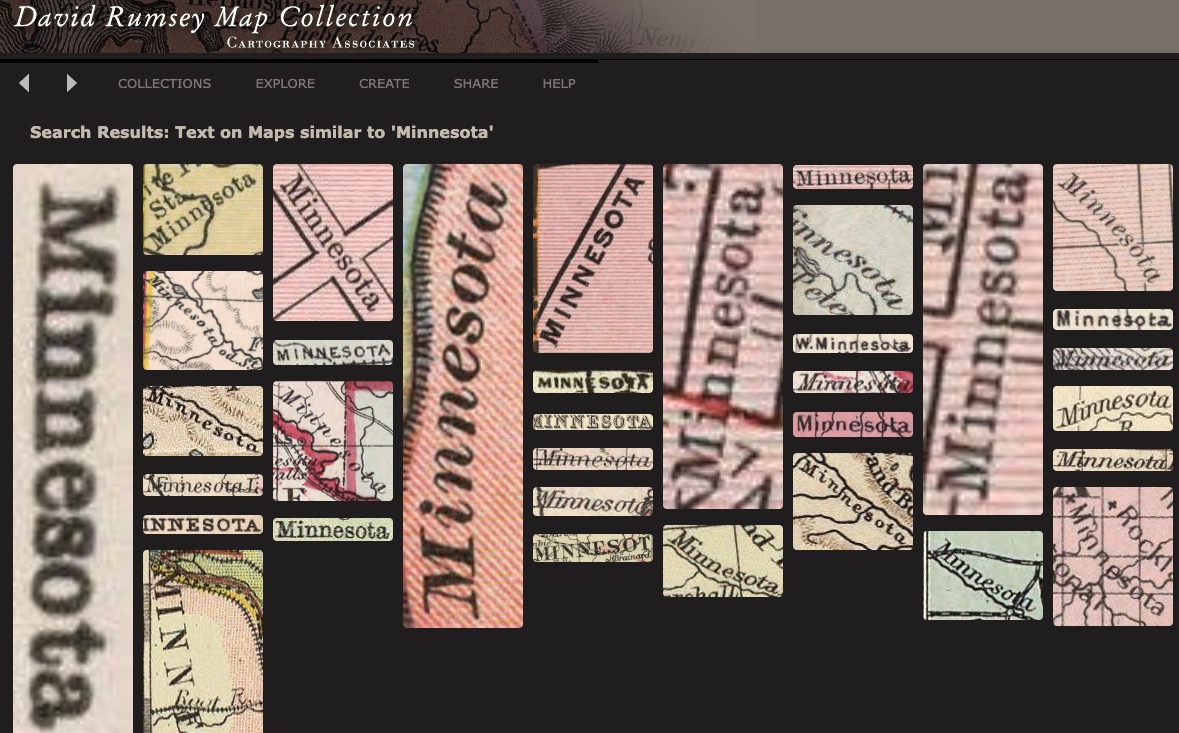}  
    \includegraphics[width=\linewidth]{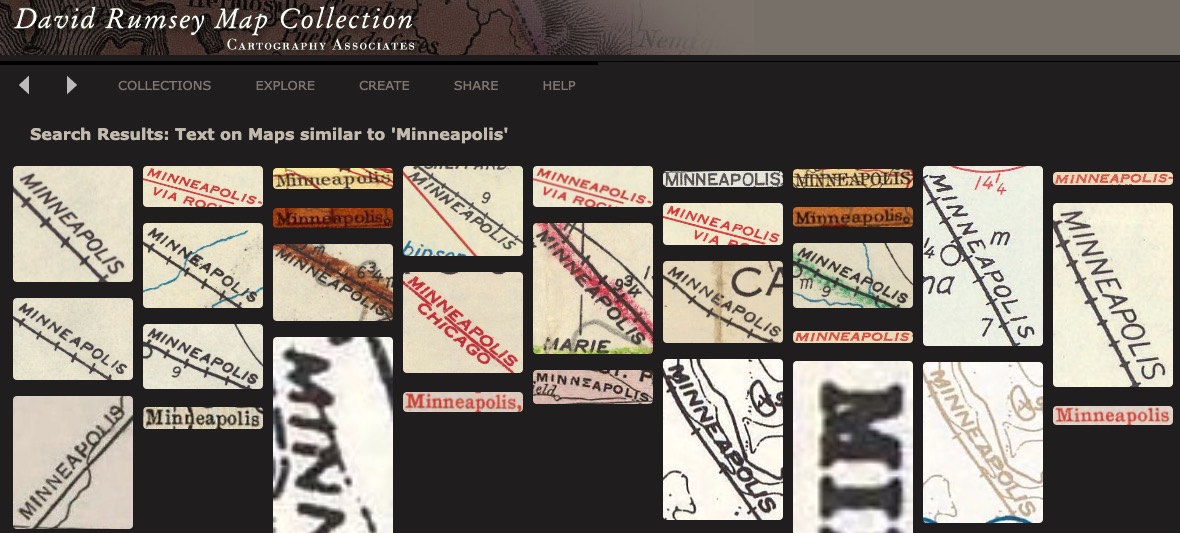}  
    \caption{The ``Search by Text-on-Maps'' interface of searching ``Minnesota'' (top) and ``Minneapolis'' (bottom). }
  \label{figure: res10}  
\end{figure}

We have deployed a version of \modelnamex trained with \synthmap, \synthtext, and human-annotated datasets in the mapKurator system~\cite{kim2023mapkurator}. The mapKurator system is a complete pipeline for efficiently extracting and linking text from large-dimension historical maps.\footnote{\url{https://github.com/knowledge-computing/mapkurator-system}} The deployment at the University of Minnesota has processed over 60,000 maps in the David Rumsey Historical Map Collection and generated more than 100 million text labels, which have been incorporated into the metadata and visualization platform by Luna Imaging that enables the ``Search by Text-on-Maps'' capability on the David Rumsey Historical Map Collection website.\footnote{\url{https://www.davidrumsey.com/home}} We provide two examples in Figure~\ref{figure: res10} of searching ``Minnesota'' (2,099 words found in the map collection) and ``Minneapolis'' (1,429 words in total). The searching function will show all the text regions and the corresponding maps containing the target word. This function supported by \modelnamex and \synthmapx largely facilitates searching through tens of thousands of historical map scans, promoting the studies such as the geographic name changes~\cite{olson2023automatic}. mapKurator with \modelnamex is also deployed at the Center for GIS, RCHSS, Academia Sinica, Taiwan and the Institute of Disaster Mitigation for Urban Cultural Heritage, Ritsumeikan University, Japan for text extraction from various types of historical maps.

\section{Related Work}
\subsection{Text Spotting on Scanned Documents} Text detection and recognition on scanned documents are important to understand the contents of documents, including scanned receipts~\cite{huang2019icdar2019}, historical newspapers~\cite{Shima2011}, manuscripts~\cite{sanchez2019set} and maps~\cite{rumsey2002david, hewitt2011map, asante2017data}. Traditional techniques usually involve careful design with heuristics for a specific document style~\cite{chiang2014survey, li2018intelligent}, e.g., separating text and background based on color~\cite{chiang-text-recognition}.
Deep neural networks have gained significant attention for automatically localizing and recognizing text from documents~\cite{li2020automatic, lombardi2020deep}. DAN is a transformer-based method for recognizing and comprehending handwritten text~\cite{coquenet2023dan}. However, the map text can be in arbitrary shapes and various styles. \citet{weinman2019deep} propose a CNN-based text detection model to localize the text and then recognize the detected text regions using CRNN~\cite{shi2016end}. However, the detection results can limit the recognition performance in two-stage approaches.

\subsection{Text Spotting on Scene Images} 
Recent end-to-end text spotting approaches mainly focus on scene images and demonstrate promising results on benchmark datasets~\cite{karatzas2015icdar, ch2020total}. Classical spotting models adopt a detect-then-recognize architecture that relies on RoI-based connectors~\cite{he2017mask, liu2018fots, liu2020abcnet, ronen2022glass} or shape transform modules~\cite{qiao2020text, wang2020all}. FOTS~\cite{liu2018fots} proposes RoIRotate to extract oriented text regions from convolutional features, while ABCNet~\cite{liu2020abcnet, liu2021abcnet2} proposes BezierAlign to handle curved text using parameterized Bezier curves. SwinTextSpotter~\cite{huang2022swintextspotter} injects the detection features into the recognition stage by predicting tight masks to suppress background noise and generate recognition features. However, these methods require careful design on RoI-based operations, which can significantly impact recognition performance. In addition, segmentation-based methods typically need heuristic grouping and post-processing~\cite{liao2020mask, wang2021pgnet, xing2019convolutional}. 


Inspired by DETR~\cite{carion2020end} and Deformable DETR~\cite{zhu2020deformable}, recent text spotters~\cite{kim2022deer, kittenplon2022towards, ye2022deepsolo, zhang2022text, ye2023deepsolo, huang2023estextspotter} explore the DETR-based framework without RoI operations and complicated postprocessing. 
TESTR~\cite{zhang2022text} adopts Deformable DETR with a dual decoder structure to refine the boundary points and recognize characters layer by layer. 
DeepSolo~\cite{ye2022deepsolo, ye2023deepsolo} proposes to sample reference points from Bezier center curves and use a single transformer decoder to predict boundary points and text. 
Although these methods show promising results on scene images, map text is typically lengthy, curved, and highly rotated. Thus, using coarse reference point(s) (e.g., the proposal center or sampled points from the center line) to sample image features can be inaccurate for the target sub-components (i.e., boundary points and characters) in a text instance. \modelnamex builds on Deformable DETR but explicitly models the boundary points and characters as the sub-components, and learn from localized image features towards the the sub-components for spotting, showing outstanding performance on map text.

Another recent direction for text detection and recognition is to incorporate the capability of language models. TrOCR~\cite{li2023trocr} crops images into small patches as tokens and inputs to a text Transformer as training LMs for text recognition. ESTextSpotter~\cite{huang2023estextspotter} introduces a vision-language communication module, enabling interaction between visual features from DETR and semantic information from a language model. UNITS~\cite{kil2023towards} addresses the challenge when having multiple types of annotations (e.g., point, polygon) by proposing a language modeling approach to model detection formats as sequences of text, improving the flexibility of adopting various types of annotations for text spotting. \modelnamex leverages the self-attention between character embeddings within a word to capture the relationships between characters, which is also inspired by the idea of the language model at the character level. 

\subsection{Synthetic Datasets for Text Spotting}
Synthetic datasets \cite{wang2012end, jaderberg2014synthetic} alleviate the need for sufficient text annotations for training text spotters. \citet{gupta2016synthetic} propose a CNN-based method to create synthetic images by blending text from various styles in existing natural scenes. \citet{liu2020abcnet} further diversifies the shapes, corpus, and styles of the text. However, these synthetic images differ significantly from historical maps regarding text and background styles. \citet{weinman2019deep} propose to dynamically synthesize images on-the-fly for training by rendering text on historical maps with noises, but the images are for recognition purpose. \citet{li2021synthetic} propose to create synthetic map images using CycleGAN to transfer a contemporary map style (e.g., OpenStreetMap) into a historical style (e.g.,  Ordnance Survey) and place text labels on them. However, this synthetic dataset contains only one background style and lacks text transcription. In contrast, \synthmapx provides numerous synthetic map images that simulate text placement and background styles of real maps, coupled with text labels (including boundary points and transcription), facilitating text spotting on historical maps.

\section{Limitations, Conclusion, and Future Work}
One limitation of \modelnamex is the use of expensive character center information. Nonetheless, we showed that the requirement can be alleviated using iterative training. This paper presented a promising end-to-end text spotting method \modelnamex and a novel synthetic map generation approach with synthetic map images, \synthmap, to detect and recognize text on historical maps. We also introduced a new benchmark dataset, \textit{Rumsey309}, for text spotting evaluation on historical maps. These resources could inspire and facilitate future research and interdisciplinary collaborations. We plan to extend our work to multilingual historical maps and group the word-level spotting results into place phrases. 


\begin{acks}
This material is based on work supported in part by the gift from David and Abby Rumsey to the University of Minnesota Foundation, the University of Minnesota (UMN), and Kartta Foundation. The authors thank the MapKurator developers for deploying \modelnamex in the system. We also thank the students from the Knowledge Lab at UMN for annotating the benchmark dataset, \textit{Rumsey309}.
\end{acks}

\bibliographystyle{ACM-Reference-Format}
\bibliography{ref}


\appendix



\section{Experimental Settings}~\label{appendix: b}

We adopt ResNet-50~\cite{he2016deep} as the backbone to extract multi-scale image features for \modelnamex and baselines. We use 6 layers in the encoder and decoder. We set the number of heads as 8 and the number of sampling points as 4 for deformable attention. We select the top 100 proposals from the encoder as queries in the decoder. The number of boundary points is 16, the maximum text length is 25, and the vocabulary size is 96. 

We train \modelnamex and baseline models on 4 NVIDIA A100 (40GB) GPUs with an image batch size of 4. The total number of iterations for pretraining is 400,000, the initial learning rate for pretraining is $1 \times 10^{-4}$, and it is decayed by a factor of 0.1 in the 300,000th iteration. We finetune the models for 10,000th iterations with a learning rate of $1 \times 10^{-5}$. We incorporate the data augmentation of random resizing and cropping during training.

\section{Effect of \synthmap}~\label{appendix: c}

Table~\ref{table: res7} shows the spotting performance of training without and with \synthmapx for the rest baselines in addition to Table~\ref{table: res4}. We observe that training with \synthmapx generally improves detection and recognition performance. However, unlike \modelnamex or TESTR, training with synthetic data only (\synthtextx + \synthmap) achieves poorer performance than \synthtextx plus human annotations in SWINTS and DeepSolo, which requires further investigation, but this indicates the necessity of training with real-world images and human annotations.

\begin{table}[htbp!]
\centering
\renewcommand{\arraystretch}{0.9} 
\caption{Effect of \synthmapx (Continue). Models are pretrained with: \synthtextx ($\mathcal{S}$), human-annotations ($\mathcal{H}$) (ICDAR, Total-Text, MLT), and \synthmapx ($\mathcal{M}$). The spotting performance are reported on \textit{Rumsey-309}.} \vspace{-0.05in}
  \begin{tabular}{c | c | c c c c} \midrule
                           & Datasets & Det-P & Det-R & Det-F & E2E-None \\ \midrule
    \multirow{3}{*}{ABCNet-v2} & $\mathcal{S}$ + $\mathcal{H}$ & 78.6 &	66.2 & 71.9 & 43.6 \\ 
                           & $\mathcal{S}$ + $\mathcal{M}$ & 78.1 & 70.0	& 73.8 & 43.8 \\ 
                           & $\mathcal{S}$ + $\mathcal{H}$ + $\mathcal{M}$ & 82.4 & 81.3 & 81.9 & 44.8 \\ \midrule
    \multirow{3}{*}{SWINTS} & $\mathcal{S}$ + $\mathcal{H}$ & 87.9	& 72.5 & 79.4 & 52.3 \\ 
                           & $\mathcal{S}$ + $\mathcal{M}$ & 76.0 & 56.2 & 64.6 & 46.1 \\ 
                           & $\mathcal{S}$ + $\mathcal{H}$ + $\mathcal{M}$ & 86.9 & 71.5 & 78.4 & 56.9 \\ \midrule
    \multirow{3}{*}{DeepSolo} & $\mathcal{S}$ + $\mathcal{H}$ & 82.7 & 64.5 & 72.5 & 56.7 \\ 
                           & $\mathcal{S}$ + $\mathcal{M}$ & 72.3	& 63.6	& 67.7 & 51.8 \\ 
                           & $\mathcal{S}$ + $\mathcal{H}$ + $\mathcal{M}$ & 78.1 & 68.1 & 72.7 & 60.1 \\ \midrule
  \end{tabular}
  \label{table: res7} 
\end{table}

\end{document}